%% file: sigconf.tex
\documentclass[author]{acmart}
\AtBeginDocument{%
  }

\setcopyright{acmlicensed}
\copyrightyear{2026}
\acmYear{2027}
\acmDOI{XXXXXXX.XXXXXXX}
\acmConference[]{(Under Review)}
\acmISBN{Under Review}

\usepackage{algorithm}
\usepackage{algpseudocodex}
\usepackage{hyperref}
\usepackage{multirow}
\usetikzlibrary{arrows.meta,calc,shapes.geometric}

\begin{document}

% \title[Fit-LRP]{Fit-LRP: Exploiting Polyhedral Constraints to Fit Affine Bounds for Non-Convex Safety Verification}
\title[Fit-LRP]{Fit-LRP: Fitting Linear Relaxations to Polytopes for Neural Network Verification over Non-Convex Regions}


\author{Brian Yarbrough}
\affiliation{%
  \institution{Purdue University}
  \city{West Lafayette}
  \state{IN}
  \country{USA}}
\email{yarbroub@purdue.edu}
\orcid{0009-0000-8493-8191}

\author{Jesse Wayment}
\affiliation{%
  \institution{Purdue University}
  \city{West Lafayette}
  \state{IN}
  \country{USA}}
\email{jwayment@purdue.edu}
\orcid{0009-0000-8493-8191}

\author{Jingbo Wang}
\affiliation{%
  \institution{Purdue University}
  \city{West Lafayette}
  \state{IN}
  \country{USA}}
\email{wang6203@purdue.edu}

\author{Philip E. Par\'{e}}
\affiliation{%
  \institution{Purdue University}
  \city{West Lafayette}
  \state{IN}
  \country{USA}}
\email{philpare@purdue.edu}

\author{Shreyas Sundaram}
\affiliation{%
  \institution{Purdue University}
  \city{West Lafayette}
  \state{IN}
  \country{USA}}
\email{sundara2@purdue.edu}

\renewcommand{\shortauthors}{Yarbrough et al.}

\begin{abstract}
State-of-the-art neural network verifiers evaluate linear relaxations over axis-aligned bounding boxes. However, verification domains in cyber-physical systems are rarely boxes. Reachability specifications, obstacle-avoidance constraints, and neural Lyapunov certificates typically yield oblique or non-convex polyhedral regions, over which box-based bounds become extremely loose.
This paper proposes Fit-LRP (Fitting Linear Relaxations to Polytopes), a framework the implements exact optimization of affine bounds over non-convex domains (represented by unions of polytopes) by solving a sequence of small, low-dimensional linear programs.
Our approach seamlessly enhances the final optimization step of LiRPA and $\alpha,\! \beta$-CROWN branch-and-bound (BaB) algorithms. 
In BaB verification, approximating oblique constraints with axis-aligned splits causes severe combinatorial blowup and risks non-termination.
By optimizing the final affine bound of a nonlinear function over the exact polyhedral state-space constraints our method eliminates box-induced over-approximations, enabling earlier search-tree pruning.
On ARCH-COMP and neural Lyapunov benchmarks, Fit-LRP reduces branching on the hardest region by $47\times$, cuts worst-case verification time by $15\times$, and with region-wise partitioning, verifies on a single CPU $6.1\times$ faster than $\alpha,\! \beta$-CROWN on a GPU.
\end{abstract}



% \received{05 November 2026}

\maketitle

\section{Introduction}\label{sec:intro}

Formally verifying properties of neural-networks is critical for proving the safe behavior of cyber-physical systems, but has been shown to be NP-hard~\cite{katz2017reluplex, bunel2020branch}.
To tackle the verification problem, state-of-the-art verifiers such as $\alpha,\! \beta$-CROWN~\cite{zhang2018efficient, wang2021beta, xu2021fast} leverage the sound affine bounds produced by auto\_LiRPA~\cite{xu2020automatic} (which uses intermediate linear relaxations of neuron activations) to execute a branch-and-bound (BaB) algorithm. While highly effective, these tools typically assume input domains are compact hyperrectangles.

In many cyber-physical applications, however, verification domains are inherently oblique and non-convex. For instance, continuous piecewise affine (CPA) neural Lyapunov functions partition the state space into polytopes that misalign with coordinate axes and are non-convex due to an origin exclusion~\cite{SamanipourTAC, kamali2025polyhedral, wayment2026hyperlyve}. Likewise, hybrid systems frequently exhibit switching guards that are arbitrary polytopes~\cite{HUANG2025101571, ARCH17:Verifying_safety_autonomous_spacecraft}.
Enforcing axis-aligned linear relaxations on such regions causes severe over-approximations of the dynamics, resulting in combinatorial BaB blowup.

Zhou et al. recently introduced \emph{Clip-and-Verify} (C\&V)~\cite{zhou2025clip} as part of the  $\alpha,\! \beta$-CROWN verifier, which then won VNN-COMP 2025~\cite{kaulen2025vnn}. C\&V rapidly prunes the BaB search space by propagating linear constraints to per-neuron relaxations. To preserve parallelism on the GPU, C\&V must approximate constraints and cannot precisely represent complex unions of polytopes. Furthermore, even if these intermediate relaxations were optimal, Salman et al.~\cite{salman2019convex} demonstrated that tightening intermediate relaxations hits fundamental limits, motivating the need to fit the final affine bound to the original input constraints during what is known as \emph{concretization}.

To bridge this gap, we observe that cyber-physical state spaces frequently have non-convex polyhedral input constraints, and we propose using these constraints to exactly fit the final affine bounds from LiRPA or $\alpha,\! \beta$-CROWN to the domain of interest. Rather than executing expensive per-neuron linear programs (LP) across intermediate layers, our approach solves a sequence of small, low-dimensional LPs to exactly optimize the final affine bound.

\paragraph{Contributions}
Our key idea is that even after computing an affine bound, optimizing the bound over an enclosing box can lose substantial precision. Fit-LRP (Fitting Linear Relaxations to Polytopes) addresses this by using light-weight LP optimization of the final affine bound over the actual polyhedral constraints.
We empirically demonstrate that our strategy eliminates box-induced looseness before branching, reducing the number of verification subproblems by up to $47\times$ and certifying cyber-physical systems 70\% faster than state-of-the-art.

\noindent All code is available at \href{https://github.com/anonymous/fit-lrp}{https://github.com/anonymous/fit-lrp}.
% \href{https://github.com/byarbrough/Fit-LRP.git}{https://github.com/byarbrough/Fit-LRP.git}.

In the remainder of this paper, Section~\ref{sec:prelim} provides preliminaries, our problem statement, and an overview of our solution. Section~\ref{sec:lp_tight} explains in detail our LP tightening approach. We present empirical results in Section~\ref{sec:experiments} before our concluding discussion in Section~\ref{sec:conclusion}.

\paragraph{Notation} We denote the set $\{1,\dots, n\}$ as $[n]$. For a vector $\mathbf{x} \in \mathbb{R}^n$, the $L_\infty$-norm is $||\mathbf{x}||_\infty = \max_i |x_i|$, $i\in [n]$. Let $\mathbf{A}^\top$ denote the transpose of a matrix $\mathbf{A}$. The notation $B|_s$ indicates that the value $B$ only holds within a specific space $s$.

\section{Problem Statement and Preliminaries}\label{sec:prelim}

This section provides preliminaries on using affine bounds to verify that a function is strictly negative on a region of interest. Then we present our problem statement, solution overview, and, finally, related work.

\subsection{Verify a Function with Affine Bounds}

We seek to verify that a nonlinear function $f$, which may include a neural network, is strictly negative on a specified region of interest. Verifying that a function is strictly negative over a region is especially of interest when determining asymptotic stability of dynamical systems~\cite{li2025two, SamanipourTAC, wang2021beta, wayment2026hyperlyve}. 
% We choose the strictly-negative property without loss of generality.
% , as a shift or negation of $f$ can encode any general property.

\paragraph{Region of interest} Let the state space $\mathcal{X} \subset \mathbb{R}^n$ be a compact polytope representing the input domain for a function $f : \mathbb{R}^n \to \mathbb{R}$.
The \emph{region of interest} $\mathcal{R} \subseteq \mathcal{X}$ is a union of $K$ convex polytopes,
\begin{equation}\label{eq:roi}
    \mathcal{R} = \bigcup_{i=1}^{K} \left\{\mathbf{x} \in \mathcal{X} \mid \mathbf{W}_i\mathbf{x} \leq \mathbf{b}_i\right\},
\end{equation}
where $\mathbf{W}_i \in \mathbb{R}^{m_i \times n}$ and $\mathbf{b}_i \in \mathbb{R}^{m_i}$ give the \emph{H-representation} of the polytope $\mathcal{P}_i$ with $m_i$ linear constraints.
In general, $\mathcal{R}$ is not axis-aligned, which we call \emph{oblique}, i.e., its facet normals are not parallel to the coordinate axes. Examples of oblique regions include the activation regions of an arbitrary ReLU controller~\cite{wayment2026hyperlyve, SamanipourTAC} or polyhedral partitions of a hybrid system~\cite{ARCH17:Verifying_safety_autonomous_spacecraft}.
For such regions, the tightest enclosing axis-aligned hyperrectangle can be arbitrarily larger than $\mathcal{R}$.
The region $\mathcal{R}$ is convex when $K=1$; a common non-convex case excludes an $L_\infty$-ball around an equilibrium at the origin.

\paragraph{Verifying the negative condition} Given a region of interest $\mathcal{R} \subseteq \mathcal{X}$ and a function $f$, we \emph{verify} that $f$ is strictly negative on $\mathcal{R}$ by proving $f(\mathbf{x}) < 0$ for all $\mathbf{x} \in \mathcal{R}$.
This verification problem is equivalent to showing that $\max_{\mathbf{x} \in \mathcal{R}} f(\mathbf{x}) < 0$, which for non-convex functions is generally NP-hard~\cite{pardalos1991quadratic}.
Note that when the state space is tiled by a union of $R$ regions,
\begin{equation}\label{eq:region_tile}
    \mathcal{X} = \bigcup_{r \in [R]} \mathcal{R}_r,
\end{equation}
then verifying that $f$ is strictly negative on every region is equivalent to verifying that $f$ is strictly negative on the entire state space:
\begin{equation}\label{eq:verify_statespace}
    \forall r \in [R], \; \forall \mathbf{x} \in \mathcal{R}_r, \; f_r(\mathbf{x}) < 0 \iff \forall \mathbf{x} \in \mathcal{X}, \; f(\mathbf{x}) < 0.
\end{equation}
\paragraph{LiRPA affine bounds} To handle the general intractability of maximizing $f$, a common approach is to prove that an upper bound on $f$ is strictly negative. 
Automatic Linear Relaxation-based Perturbation Analysis for Neural Networks (auto\_LiRPA)~\cite{xu2020automatic} uses linear relaxations to compute certified affine upper (and lower) bounds of a function.
LiRPA composes \emph{intermediate} bounds on individual neurons into a \emph{final} bound. In this work we are concerned primarily with the final upper bound. 
The upper bound of $f$ corresponds to a hyperplane expressed as a matrix $\overline{\mathbf{A}}$ and offset $\overline{\mathbf{c}}$, and holds within a given axis-aligned hyperrectangular box $\mathcal{B}$:
\begin{equation}\label{eq:lirpa_box}
    f(\mathbf{x}) \leq \overline{\mathbf{A}}\mathbf{x} + \overline{\mathbf{c}} \quad \forall \mathbf{x} \in \mathcal{B} \subseteq \mathcal{X}.
\end{equation}

\paragraph{Concretization} Suppose~\eqref{eq:lirpa_box} gives an upper bound on $f$ for a box that intersects a region: $\mathcal{R} \cap \mathcal{B} \subseteq \mathcal{X}$.
The next step towards proving $f$ is strictly negative on $\mathcal{R}$ is to \emph{concretize} the maximum value this bound can obtain relative to the region of interest $\mathcal{R}$. The maximum concrete value on $\mathcal{R}$ might be much smaller than the maximum on $\mathcal{B}$, especially when $\mathcal{R}$ is oblique or non-convex, because~\eqref{eq:lirpa_box} must account for the behavior of $f$ over a box that is potentially much larger than the actual region of interest.
An exact optimization over $\mathcal{R}$ can prove expensive or be difficult to parallelize on the GPU effectively, so the constraints of $\mathcal{R}$ may be relaxed to a convex over-approximation, $\hat{\mathcal{R}}$. Relaxing the constraints in this way is still sound, but will yield a looser bound. Concretization is then
\begin{equation}\label{eq:lirpa_concrete}
    \begin{aligned}
    y = \max_{\mathbf{x}} \quad & \overline{\mathbf{A}}\mathbf{x} + \overline{\mathbf{c}}, \quad & \mathbf{x} \in \hat{\mathcal{R}} \cap \mathcal{B}.
\end{aligned}
\end{equation}
It holds that $f(\mathbf{x}) \leq y$, for all $\mathbf{x} \in \hat{\mathcal{R}} \cap \mathcal{B}$ and $\mathcal{R} \subseteq \hat{\mathcal{R}}$, so if~\eqref{eq:lirpa_concrete} yields $y < 0$, then $f(\mathbf{x}) < 0$ is verified for all $\mathbf{x} \in \mathcal{R} \cap \mathcal{B}$.
Conversely, if $y \geq 0$, it does not necessarily mean verification fails; it often indicates the bound is simply too loose.

\paragraph{Branch-and-Bound (BaB)} If the optimum of~\eqref{eq:lirpa_concrete} is non-negative but the true function value at that optimizer, $\mathbf{x}^*$, is strictly negative ($y \ge 0$ while $f(\mathbf{x}^*) < 0$), verification is inconclusive and a tighter bound is required.
A well-known method to tighten bounds is \emph{branch-and-bound} (BaB), which splits an inconclusive box into smaller subdomains ($\mathcal{B} = \mathcal{B}_1 \cup \mathcal{B}_2$) so that tighter bounds might be obtained separately on each box. Although BaB is a sound method, its search space can explode exponentially; tighter bounds mitigate branch explosion by allowing earlier pruning, but computing tighter bounds costs more per node~\cite{bunel2020branch, ladner2023automatic, zhou2025clip}.
Tightening intermediate bounds on hidden network layers quickly yields diminishing returns~\cite{salman2019convex}, so a cheap approximation of input constraints is typically superior to exact evaluation~\cite{zhou2025clip}. However, we shall show that concretizing the final affine bound~\eqref{eq:lirpa_concrete} against a function's \emph{exact} polyhedral input constraints (the facets of $\mathcal{R}$) outperforms approximating the constraints because the LP optimization is cheap while the reward is high: exact concretization drastically reduces the number of branches required to verify that $f$ is strictly negative on $\mathcal{R}$, especially when the property holds by small margins.

\subsubsection{Clip-and-Verify}
Our approach can directly augment the Clip-and-Verify (C\&V) method introduced by Zhou et al. in~\cite{zhou2025clip}.
C\&V is integrated into the $\alpha,\! \beta$-CROWN verifier and provides enhanced BaB performance by optimizing intermediate bounds via linear constraints.
Specifically, their ``complete'' clipping method observes that if $\mathcal{R}$ only imposes a single additional linear constraint on $\mathcal{B}$ then concretizing~\eqref{eq:lirpa_concrete} can be solved exactly by formulating the Lagrangian dual of the optimization problem.
In the case of multiple linear constraints, Zhou et al. show that sequentially optimizing the Lagrangian dual of each constraint is more efficient for large-scale verification problems than simultaneously optimizing under all constraints, even though it produces an over-approximation of the true feasible region. In their experiments, C\&V optimization resulted in up to a 96\% reduction in the number of BaB subproblems required for $\alpha,\! \beta$-CROWN to verify network properties.
Their demonstration shows that using C\&V to tighten per-neuron bounds in intermediate layers of a neural network is superior to bound-tightening via linear programming (LP). However, their emphasis is on per-neuron optimization rather than concretization of the final affine bound. Furthermore, their approach is unable to clip to non-convex regions. Our solution enhances C\&V for $\mathcal{R}$ with $K>1$ via exact concretization of the final upper bound.

\subsection{Problem Statement and Solution}

\begin{figure}
\centering
\input{space_tikz}
\caption{Position cut of the spacecraft rendezvous (Section~\ref{sec:spre}) region $\mathcal{R}$. The octagonal switching guard is represented via affine constraints (solid black). Removing the docking cube (red, enlarged) makes $\mathcal{R}$ non-convex and forms the union of $K=4$ disjoint convex tiles $\mathcal{P}_i$ (dashed) in a pinwheel configuration, where each tile corresponds to $\mathcal{R}$ intersected with two half-planes. The tightest axis-aligned root bounding box $\mathcal{B}_0$ (blue) over approximates the diagonal faces of $\mathcal{R}$.
For an axis-aligned velocity facet (green, $\nabla\overline{A^\prime}$ along $\vec{x}$), box-based concretization is already exact ($\hat{y}^\prime = y^\prime$). However, for a diagonal facet (purple), box concretization yields $\hat{y}_0 = +23.1$ (diamond) at a corner outside $\mathcal{R}$, which fails to verify.
Conversely, Fit-LRP computes exact values $z_i$ (dots), taking the maximum $y = z_1 < \hat{y}_0$ on the octagon facet (orange) in $\mathcal{P}_1$; here, $y=-0.82$ and the facet is verified.
}
\Description[Position cut of the spacecraft rendezvous region with octagonal guard and pinwheel tiles]{A 2D diagram illustrating a position cut of the spacecraft rendezvous region $\mathcal{R}$. The outer boundary is an octagon representing a switching guard, surrounding a central red square representing an enlarged docking box. The non-convex region outside the docking box is partitioned into four pinwheel tiles ($\mathcal{P}_1$ through $\mathcal{P}_4$) via dashed boundaries. A blue bounding box $\mathcal{B}_0$ encloses the octagon. Annotated elements include gradient vectors ($\nabla\overline{\mathbf{A}}$ and $\nabla\overline{\mathbf{A}'}$), a diagonal level set, a conservative box maximizer point $\hat{y}_0$ (diamond), and per-tile LP concretization points $z_1$ through $z_4$ (dots) along an active octagon facet, demonstrating how per-tile linear programming tightens verification bounds compared to box-based concretization.}
\label{fig:space}
\end{figure}
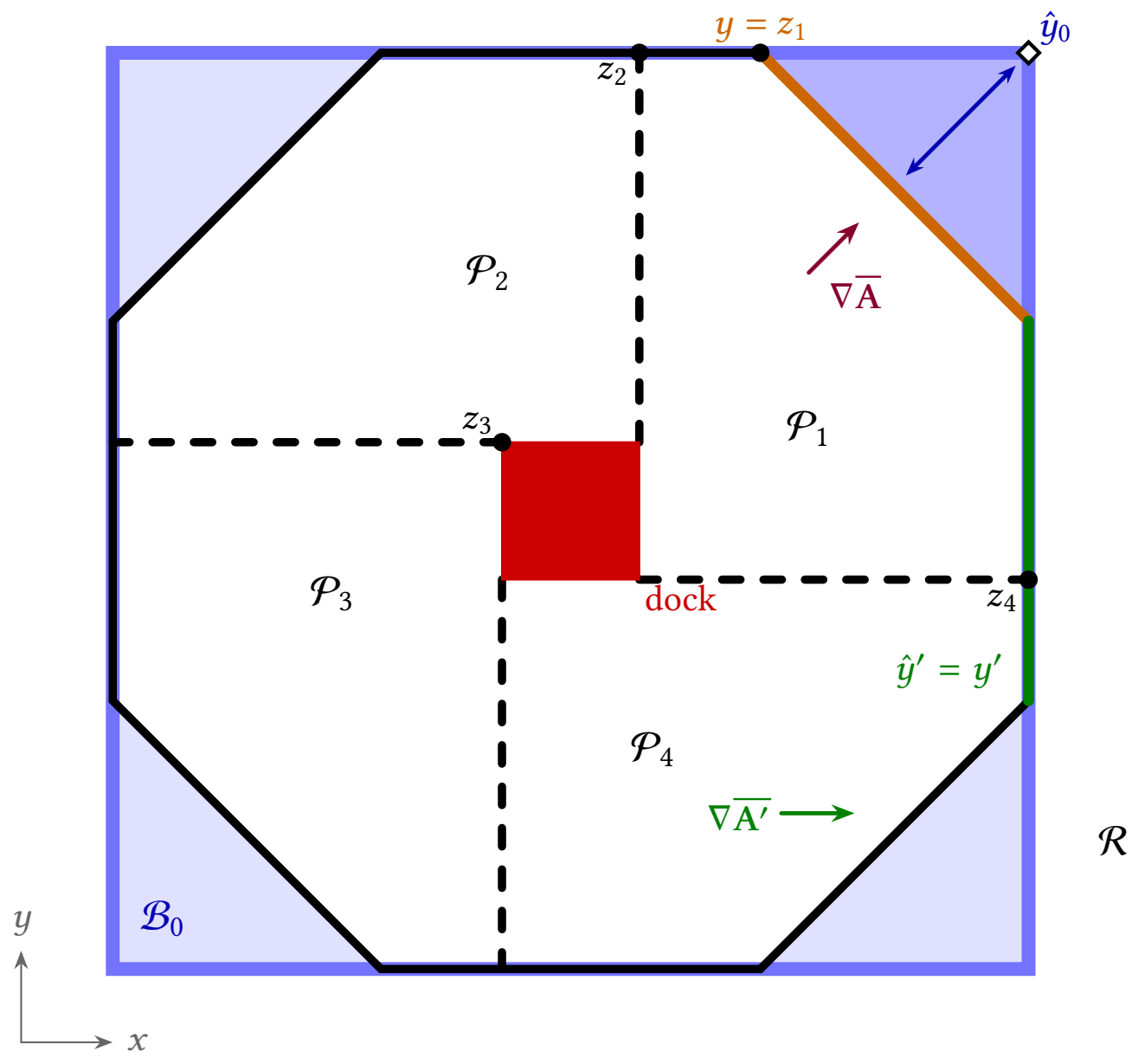

\paragraph{Problem statement} Given a possibly non-convex region of interest $\mathcal{R}$ defined by~\eqref{eq:roi} as a union of polyhedra, mitigate BaB explosion when verifying the function $f$ is strictly negative on $\mathcal{R}$ by tightening the bound from~\eqref{eq:lirpa_box}.

\paragraph{Solution: Fit-LRP} Our fix is simple: concretize~\eqref{eq:lirpa_concrete} by solving a series of $K$ linear programs and taking the maximum of their solutions as $y$. That is:
\begin{equation}\label{eq:sol_lp}
    \begin{aligned}
    z_i = \max_{\mathbf{x}} \quad & \overline{\mathbf{A}}\mathbf{x} + \overline{\mathbf{c}} \\
    \text{s.t.} \quad & \mathbf{x} \in \mathcal{P}_i \cap \mathcal{B},
\end{aligned}
\end{equation}
\begin{equation}\label{eq:sol_max}
    y := \max_{i \in [K]} z_i.
\end{equation}
The optimal point $\mathbf{x}^*$ is the maximizer corresponding to $y$:
\begin{align}
        \mathbf{x}_i^* =& \arg \max_\mathbf{x} \overline{\mathbf{A}}\mathbf{x} + \overline{\mathbf{c}},\\
        \mathbf{x}^* :=& \arg \max_{i\in[K]} z_i.
\end{align}
Put differently, the result of~\eqref{eq:sol_lp} and~\eqref{eq:sol_max} is an exact solution to~\eqref{eq:lirpa_concrete}:
\begin{equation}\label{eq:y_max}
    y = \max_\mathbf{x} \overline{\mathbf{A}}\mathbf{x} + \overline{\mathbf{c}}, \quad \mathbf{x} \in \mathcal{R} \cap \mathcal{B},
\end{equation}
which implies
\begin{equation}
    f(\mathbf{x}) \leq y, \quad \mathbf{x} \in \mathcal{R} \cap \mathcal{B}.
\end{equation}

When $\mathcal{P}_i \cap \mathcal{B} = \emptyset$, the LP is infeasible, so we set $z_i = -\infty$, and if all LPs are infeasible then $\mathcal{R}\cap \mathcal{B}=\emptyset$ and we prune the branch immediately. 
Solving each LP is fast and subsequent LPs can be warm-started from the dual-simplex bases found while solving an adjacent polytope.
It is imperative to note that we only exactly solve $K$ optimization problems on the final affine bound for each branch, which is far fewer than the number of optimization problems required for intermediate bounding of neurons.

We refer to our solution as \emph{Fit-LRP} (Fitting Linear Relaxations to Polytopes). We integrate our solution with into LiRPA, but the technique applies to any affine lower or upper bound of a function, regardless of which tool generates the bound. Additionally, Fit-LRP could be easily extended to any convex-bound framework by replacing the LPs with an appropriate optimizer.

\subsubsection{Motivational Example}
Figure~\ref{fig:space} shows the verification of a velocity-limit specification adapted from the ARCH-COMP nonlinear hybrid systems ``Spacecraft Rendezvous'' benchmark~\cite{ARCH17:Verifying_safety_autonomous_spacecraft} (Section~\ref{sec:benchmak_sys}).
During close-range rendezvous, the region $\mathcal{R} \in \mathbb{R}^4$ is defined by an octagonal position guard (pictured) crossed with an octagonal velocity guard; the region is non-convex due to the exclusion of a cubic dock. Adherence to the velocity limit is verified by showing a barrier function $f$ is strictly negative on $\mathcal{R}$.
LiRPA calculates the linear upper bound $\overline{\mathbf{A}}$ on $f$ for all $\mathbf{x} \in \mathcal{B}_0$, the tightest box containing $\mathcal{R}$.
When concretizing~\eqref{eq:lirpa_concrete}, if the gradient $\nabla \overline{\mathbf{A}}$ is normal to an axis-aligned facet of the octagon, then LiRPA concretizes the optimal value, $\hat{y} = y$. However, if $\nabla \overline{\mathbf{A}}$ is directed towards a diagonal of the octagon, LiRPA concretizes a conservative approximation; meanwhile, our LP-tightening technique concretizes the optimal value, so $f(\mathbf{x}) \leq y < \hat{y}_0$ for all $\mathbf{x} \in \mathcal{R} \cap \mathcal{B}_0$.
As we observed in our experiments (Section~\ref{sec:results}), the difference between these two concrete values can be the difference between Fit-LRP verifying the condition immediately versus splitting $\mathcal{B}_0$ into subproblems.

\section{Fit-LRP}\label{sec:lp_tight}

In this section, we describe how Fit-LRP uses the max of a series of linear programs to concretize~\eqref{eq:lirpa_concrete} and can be applied directly to LiRPA in an elementary branch-and-bound procedure. We also describe how to augment Clip-and-Verify with our approach.

\subsection{Fit-LRP BaB}

Algorithm~\ref{alg:lirpa_bab} presents a deliberately simple branch-and-bound (BaB) scheme. The bisection of inconclusive boxes in line~\ref{alg:lirpa_bab:bisect} merely splits the box along the largest axis~\cite{bunel2020branch}.
We chose an un-optimized branching approach to highlight the benefit of exactly concretizing~\eqref{eq:lirpa_concrete}. Obtaining the true maximum constrained to the region of interest reduces the number of subproblems that must be solved because the tighter bound reduces the number of required branches.

\begin{algorithm}[h]
\caption{Fit-LRP BaB}
\label{alg:lirpa_bab}
\begin{algorithmic}[1]
    \Require Function $f(\mathbf{x})$, region of interest $\mathcal{R} \subseteq \mathcal{X}$ defined as union of $K$ polytopes. Iteration budget $I_{\max}$.
    \Ensure $f$ is strictly negative on $\mathcal{R}$ or counterexample $f(\mathbf{x}^*) \geq 0$.
        \State $\mathcal{B}_0 \gets \Call{BoundingBox}{\mathcal{R}}$
        \State Initialize queue $\mathcal{Q}\gets\{\mathcal{B}_0\}$.
    \While{$\mathcal{Q}\neq\emptyset$ and processed boxes $<I_{\max}$}
        \State Remove a box $\mathcal{B}$ from $\mathcal{Q}$.
        \State $\overline{\mathbf{A}},\ \overline{\mathbf{c}} \gets \Call{LiRPA}{f(\mathbf{x}), \mathcal{B}}$ \Comment{Eq.~\eqref{eq:lirpa_box}}
        \For{$i \in [K]$}\Comment{Eq.~\eqref{eq:roi}}
            \State $z_i \gets \max_{\mathbf{x}} \; \overline{\mathbf{A}}\mathbf{x} + \overline{\mathbf{c}} \quad \text{s.t.} \quad \mathbf{x} \in \mathcal{P}_i \cap \mathcal{B}$ \Comment{LP~\eqref{eq:sol_lp}}
        \EndFor
        
       \If{all LPs infeasible} \State \textbf{continue} \Comment{$\mathcal{B} \cap \mathcal{R} = \emptyset$, branch pruned.} \EndIf     
       \State $y,\ \mathbf{x}^* \gets \arg\max_i z_i$\Comment{Solving~\eqref{eq:sol_max} implies~\eqref{eq:y_max}}
        \If{$y < 0$}
            \State \textbf{continue} \Comment{$f$ is verified negative in this branch}
        \EndIf
        \If{$f(\mathbf{x}^*) \ge0$}
            \State \Return ($\mathbf{x}^*$) \Comment{Counterexample, $f$ fails verification}
        \EndIf
        \State $\mathcal{B}_1,\ \mathcal{B}_2 \gets \Call{Bisect}{\mathcal{B}}$\label{alg:lirpa_bab:bisect} \Comment{Bound was inconclusive}
        \State Insert $\mathcal{B}_1,\ \mathcal{B}_2$ into $\mathcal{Q}$.
    \EndWhile
    \If{$\mathcal{Q}=\emptyset$}
        \State \Return $\emptyset$ \Comment{$f$ is verified strictly negative on $\mathcal{R}$}
    \Else
        \State \Return \textsc{Unknown} \Comment{Reached $I_{\max}$}
    \EndIf
\end{algorithmic}
\end{algorithm}

\subsection{Clip-and-Verify with Fit-LRP}\label{sec:lp_tight:cv}

To demonstrate our method within a more sophisticated branch-and-bound scheme, we integrate our LP-based concretization into $\alpha,\!\beta$-CROWN's Clip-and-Verify (C\&V) algorithm. 
While C\&V relies on standard relaxations for efficient \emph{intermediate neuron-bound tightening}, its final concretization over each subdomain polytope $\mathcal{P}_i$ typically relies on sequential dual optimization. 
Fit-LRP complements C\&V by replacing this final step with our exact LP solution~\eqref{eq:sol_lp}. 
Because optimizing over the polytope is restricted to the final concretization step rather than propagated through intermediate layers, it remains highly inexpensive while avoiding relaxation slack. 
Furthermore, whereas standard C\&V delays clipping constraints until after the first split, Fit-LRP applies this exact final-bound optimization at the root box, enabling immediate region verification prior to branching. We refer to this configuration as C\&V, Fit-LRP.

\section{Experiments}\label{sec:experiments}

Here we present the empirical results of our solution on several dynamical systems. Every problem seeks to verify that $f(\mathbf{x}) <0$ for all $\mathbf{x} \in \mathcal{X}$. The regions comprising each state space are oblique and, in some cases, non-convex. We executed experiments on AMD EPYC 7763 Milan 64-core CPU and an AMD M120 GPU. We used Claude Code~\cite{claudecode} to facilitate experiments, check our code, and create figures. Complete code, data, and SLURM artifacts are available.

\subsection{Benchmark Systems}\label{sec:benchmak_sys}
We benchmark a hybrid system from ARCH-COMP and two CPA neural Lyapunov functions.

\subsubsection{ARCH-COMP Spacecraft Rendezvous (SPRE)~\cite{ARCH17:Verifying_safety_autonomous_spacecraft}}\label{sec:spre}
We evaluate our approach on the ARCH-COMP SPRE velocity benchmark, which models planar relative spacecraft motion under an LQR controller.
Safety requires proving velocity limits ($\|v\| \le 3.3\text{ m/min}$) over the non-convex region $\mathcal{R} = (O_p \times O_v) \setminus D$, where $O_p$ and $O_v$ are octagonal position/velocity guards, and $D$ excludes the central $0.2\text{ m}$ docking box.
For each of the $8$ velocity facet constraints $a_j^\top v \le v_{\max}$ ($v_{\max} = 3.049\text{ m/min}$) with closed-loop rate $\lambda = 19.2$, this yields the barrier verification query:
\begin{equation}
    f_j(z) = a_j^\top \dot{v}(z) + \lambda (a_j^\top v - v_{\max}) < 0, \quad \forall z \in \mathcal{R}.
\end{equation}
To handle the non-convexity, we tile $\mathcal{R}$ using a pinwheel pattern of $K=4$ disjoint polytopes (Figure~\ref{fig:space}).

\subsubsection{CPA Neural Lyapunov Functions}
We certify asymptotic stability using continuous-piecewise affine (CPA) neural Lyapunov functions. 
A valid Lyapunov function requires a negative Lie derivative, $\dot{V}(\mathbf{x}) = \nabla V^\top f(\mathbf{x}) < 0$, for all $\mathbf{x} \in \mathcal{X} \setminus \mathbf{0}$.
Following~\cite{wayment2026hyperlyve}, we partition the state space $\mathcal{X}$ into $R$ polyhedral affine regions defined by ReLU activation patterns and evaluate the Lie decrease condition independently per region.
Excluding a small neighborhood about the origin ($\|\mathbf{x}\|_\infty < h$ for $h=10^{-6}$) makes intersecting regions non-convex ($K>1$), while all other regions remain convex ($K=1$). 
This yields the per-region verification problem:
\begin{equation}\label{eq:lyap_verify}
    \nabla V|_{\mathcal{R}_r}^\top f(\mathbf{x}) < 0, \quad \forall r \in [R], \; \forall \mathbf{x} \in \mathcal{R}_r \text{ s.t. } \|\mathbf{x}\|_\infty \ge h.
\end{equation}

\paragraph{Two-Link Planar Robot Arm (TLPRA)~\cite{sun2021learning}}
The fully actuated two-link arm has four states $(q_1, q_2, \dot q_1, \dot q_2)$, two torque inputs, $m_1 = 0.8$, $m_2 = 2.3$ and unit link lengths.
A \texttt{tanh} controller with 4 hidden layers of 64 neurons regulates the arm to the origin, with gravity-compensating torque $\mathbf{u}^*$ there.
Since $\det \mathbf{M}(\mathbf{q}) \geq 1.84 > 0$, we certify the sign-equivalent, division-free field $\det \mathbf{M}(\mathbf{q})\,\dot{\mathbf{x}}$.
On $\mathcal{X} = [-\pi/8, \pi/8]^4$, a Lyapunov network with 3 hidden layers of 14 neurons has 3,068 regions, $756$ non-convex.

\paragraph{Mitigated Epidemic (MSIS)~\cite{lajmanovich1976deterministic}}
A neural network controller with 4 layers of 16 \texttt{tanh} neurons mitigates a susceptible, infected, susceptible (SIS) networked epidemic spreading over a graph. We chose a connected graph of 8 nodes, 10 edges, and a reproduction number $R_0 = 1.07$, implying the epidemic will not die out without any mitigation intervention. On $\mathcal{X} =[0, 1]^8$, a Lyapunov network with 3 hidden layers of 12 neurons has 2,710 regions, 218 non-convex.

\subsection{Compared BaB Algorithms}
We compare three approaches that verify the left-hand side of~\eqref{eq:verify_statespace} by tiling $\mathcal{X}$ with~\eqref{eq:region_tile} and verifying each of the $R$ regions independently.
\begin{itemize}
    \item \textbf{C\&V, $\mathcal{R}$}: $\alpha,\! \beta$-CROWN with complete Clip-and-Verify~\cite{zhou2025clip}. This is the baseline. It concretizes every bound by single-pass dual coordinate ascent over the approximate clip constraints.
    \item \textbf{C\&V, Fit-LRP}: identical to C\&V except we concretize the final bound with LP~\eqref{eq:sol_lp} and maximization~\eqref{eq:sol_max} (Section~\ref{sec:lp_tight:cv}).
    \item \textbf{Fit-LRP BaB}: Algorithm~\ref{alg:lirpa_bab}, LiRPA bounds with Fit-LRP concretization in an elementary branch-and-bound approach.
\end{itemize}
We also baseline the right-hand side of~\eqref{eq:verify_statespace} as a single problem:
\begin{itemize}
    \item \textbf{C\&V, GPU}: $\alpha,\! \beta$-CROWN with complete C\&V, on a GPU.
\end{itemize}
All approaches branch based on the widest input dimension of the current box, so differences in branch counts come from how tight the bounds are, not from the branching heuristic.

\subsection{Results}\label{sec:results}

\begin{table}[t]
    \centering
    \small
    \caption{Comparison of branching and time to verify systems.
    Under each system is the number of regions (non-convex in parentheses).
    Branches are shown as the total for the system and the maximum for any single region. Time is measured as \emph{work} ($T_1$), the time to verify on a single processor, and \emph{span} ($T_\infty$), the theoretical time to solve on infinite parallel CPUs, captured as the time to solve the slowest region.
    C\&V, GPU is one specification over the whole domain.
    For each system the top configurations are baseline, while the bottom two configurations enhance Clip-and-Verify with Fit-LRP and use Fit-LRP directly on LiRPA in a simple branch-and-bound (Algorithm~\ref{alg:lirpa_bab}), respectively.
    Our LP-optimization of the final affine bound reduces the number of branches and time required to solve each system, saving millions of subproblems and making a single CPU $T_1$ over 7-hours faster than a GPU.}
    \label{tab:results}
    \input{results}
\end{table}

Table~\ref{tab:results} shows that Fit-LRP reduces the number of branches and the time required to verify each system. In particular, enhancing C\&V with Fit-LRP decreases the total number of branches by 23\%-71\%. 
Furthermore, our method verifies all 8 SPRE regions without requiring any branching, unlike C\&V, $\mathcal{R}$, which must approximate the polytopes tiling the octagonal region (Figure~\ref{fig:space}), resulting in loose concretization triggering 6 branches each.
We measure the total \emph{work} to verify a system as the time it would take to solve on a single CPU core, denoted $T_1$.
We denote \emph{span}, the theoretical limit to verify the system on infinite parallel CPUs, as $T_\infty$ and measure it as the time required to verify the slowest region in the system.
Fit-LRP methods must solve additional LPs, but the span for any Fit-LRP method was always smaller than those without it.

\begin{figure}
    \centering
    \includegraphics[width=\columnwidth]{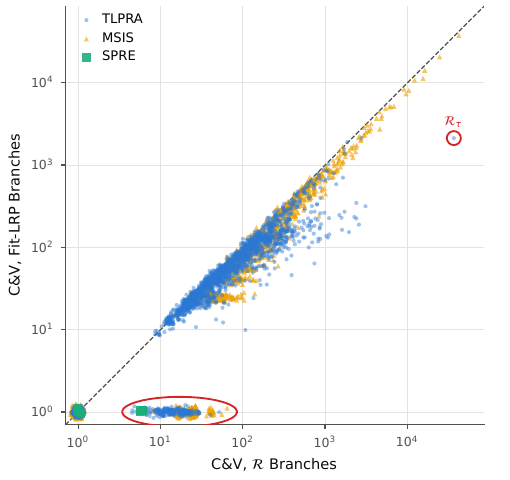}
    \Description[Log-log scatter plot comparing per-region branching of C\&V, $\mathcal{R}$ against C\%V, Fit-LRP]{A log-log scatter plot comparing the number of per-region branches required by C&V, $\mathcal{R}$ (x-axis) versus C&V, Fit-LRP (y-axis) across three benchmarks: TLPRA (blue circles), IPNN (orange triangles), and SPRE velocity (green squares). A diagonal dashed line represents equal branching ($y = x$). Nearly every point lies on or below the diagonal: C&V, Fit-LRP needed fewer branches on 52\% of regions, the same number on 48\%, and more on only 10 of 3,137 regions. A horizontal band of points along the bottom of the plot, at one branch on the y-axis and between 1 and about 50 branches on the x-axis, shows regions that C&V, Fit-LRP verified without branching while C&V, $\mathcal{R}$ had to branch. A single isolated point at the upper right, circled in red and labeled $\mathcal{R}_\tau$, is the slowest region: 36,372 branches with C&V, $\mathcal{R}$ versus 1,890 with C&V, Fit-LRP, where solve time drops from 103 seconds to 10 seconds.}
    \caption{Comparing per-region branching of C\&V, $\mathcal{R}$ against C\&V, Fit-LRP.
    Most regions are below the diagonal dashed line $y = x$, indicating that C\&V, $\mathcal{R}$ needed more branches to verify a particular region than C\&V, Fit-LRP did.
    The cluster (red circle) around coordinates (10, 1) is 737 of 5,786 regions (12.7\%): 498 MSIS, 235 TLPRA and 4 SPRE. These regions were verified by Fit-LRP  on the first bound without any branching, while C\&V, $\mathcal{R}$ had to branch 5-57 times.
    The slowest region, denoted $\mathcal{R}_\tau$ (red circle), required 36,372 branches with C\&V, $\mathcal{R}$, whereas Fit-LRP required only 1,890 branches; fewer branches decreased solve-time for $\mathcal{R_\tau}$ from over 1.7 minutes to 10.4 seconds.}
    \label{fig:cvlp}
\end{figure}

\paragraph{Zero-Split Regions} The red-circled cluster at coordinates (10, 1) in Figure~\ref{fig:cvlp} was certified by Fit-LRP on the first bound without any bounding-box splits, whereas C\&V, $\mathcal{R}$ required multiple branches.
All eight SPRE regions (green) were certified on the first bound by Fit-LRP, as indicated by the 1 in the ``max'' column of Table~\ref{tab:results}. For the four regions with diagonal facets in Figure~\ref{fig:space}, C\&V concretizes $+23.1$ at the corner of the box bound, requiring 5 additional branches before verifying; meanwhile, Fit-LRP's exact LP concretizes $-0.82$, which immediately verifies the region without any splitting.
This result highlights how our method of optimizing against the exact input constraints can completely eliminate branching where C\&V's approximate constraint propagation is insufficient.

\paragraph{TLPRA Long-Tail Region} Figure~\ref{fig:cvlp} highlights region $\mathcal{R}_{\tau}$, which required over 36-thousand branches for C\&V, $\mathcal{R}$ to verify and had $T_\infty = 103$s.
Although $\mathcal{R}_{\tau}$ is convex, it is relatively large, with many constraints. It has 11 oblique facets, which places it in the 90th percentile of TLPRA regions, and a 77th-percentile volume.
Meanwhile, $\mathcal{R}_\tau$ has the smallest Lie decrease margin of any region in the system: $\max_{\mathbf{x} \in \mathcal{R}_{\tau}} \nabla V|_{\mathcal{R}_{\tau}}^\top f(\mathbf{x}) = -5.3\times10^{-5}$.
Although large volumes already loosen the upper bounds on the initial box $\mathcal{B}$, C\&V's concretization is $1.5\times$ looser than Fit-LRP. Recall that C\&V sequentially applies linear constraints with increasing inexactness, so during branching concretization with 11 facets is significantly looser than with Fit-LRP.
As a result, $\mathcal{R}_{\tau}$ requires $13\times$ more branches than the next hardest region.
Enhancing C\&V with Fit-LRP decreased the required number of branches by $95\%$, lowering the verification time for $\mathcal{R}_\tau$ to just 10.4s.
 (Note that Table~\ref{tab:results} shows $T_\infty = 12$s because a different region is slower than $\mathcal{R}_\tau$ when Fit-LRP is applied.)
Compounding similar improvements across all 3,068 regions decreases $T_1$ for C\&V, Fit-LRP by nearly 15 minutes.

\paragraph{GPU vs.\ CPU} Typically, $\alpha,\!\beta$-CROWN is deployed on a GPU, but Fit-LRP BaB verifies each system faster on a single CPU core than C\&V does on an enterprise-grade GPU: SPRE $2.5\times$, TLPRA $3.3\times$ (Table~\ref{tab:results}). On MSIS, the GPU search exceeded $127$ million branches and takes 7 hours, while the Fit-LRP BaB verified with only 38 thousand branches and $T_1=69$ minutes. More importantly, when we parallelized region verification on a 64-core CPU, $T_{64}=81$ seconds ($317\times$ faster than GPU); with additional cores $T_\infty=4.8$ seconds could be approached.
Region-wise partitioning provides much of this advantage. For example, on TLPRA C\&V, $\mathcal{R}$ alone is $2.8\times$ faster than the GPU, and Fit-LRP extends the gap to $3.3\times$.
The whole-domain BaB search is vastly larger, requiring millions more branches.
We attribute this to 1) refuting the disjunction of $2n$ clauses that excludes the neighborhood of the origin, and 2) for CPA Lyapunov functions, relaxing the piecewise-constant gradient $\nabla V$ jointly with the closed-loop dynamics, whereas per-region methods use the analytic $\nabla V$ and only relax the dynamics.

\section{Future Work and Conclusion}\label{sec:conclusion}

We demonstrated that exact optimization of a nonlinear function's final affine bound drastically reduces the subproblems and time needed to verify cyber-physical safety properties. Integrating with existing intermediate bound tightening and pruning methods, our CPU-based exact approach can outperform state-of-the-art GPU BaB on challenging long-tail instances. 

This work opens several research directions. First, since linear programming is CPU-centric, our current solution lacks GPU support. Reconciling the tension between massively parallel GPU intermediate bounds and CPU-based final affine bounds to enable parallel GPU execution is vital for scaling to larger systems.
Second, optimal domain partitioning into polytopes remains challenging due to the complexity of enumerating activation regions and handling diverse hybrid system tilings. Future work will focus on developing automated techniques to construct verification-optimized regions.

\begin{acks}
This work is supported in part by the US National Science Foundation (NSF-ECCS \#2238388, NSF CNS-2333487, NSF-CMMI \#2535574).
\end{acks}

\bibliographystyle{ACM-Reference-Format}
\bibliography{ref}

\end{document}

%% file: space_tikz.tex
% % Requires in preamble: \usepackage{tikz,amsmath,amssymb,xcolor}
\begin{tikzpicture}[scale=0.7, line cap=round,
    hull/.style   ={black, line width=1.8pt},
    bbox/.style   ={blue!55, line width=3pt, line join=miter},
    poly/.style   ={black, dash pattern=on 4pt off 5pt, line width=1.8pt},
    guard/.style  ={black!55, dash pattern=on 1pt off 2pt, line width=0.8pt},
    pillar/.style ={red!80!black, fill=red!80!black},
    level/.style  ={orange!80!black, densely dotted, line width=1.1pt},
    axisc/.style  ={green!50!black},
    single/.style ={draw=black, fill=white, line width=0.8pt, minimum size=5pt,
                    inner sep=0pt, diamond},
    piece/.style  ={fill=black, circle, inner sep=0pt, minimum size=4pt}]

  \def\a{5}      % octagon offset a_p = 100 m
  \def\s{2.071}  % half facet length 100 tan(22.5 deg) m
  \def\h{0.75}   % docking box half-width (enlarged)

  % Inscribed octagon O_p: facet normals at k*45 deg, vertices at 22.5 + k*45 deg
  \coordinate (V1) at ( \a, \s);  \coordinate (V2) at ( \s, \a);
  \coordinate (V3) at (-\s, \a);  \coordinate (V4) at (-\a, \s);
  \coordinate (V5) at (-\a,-\s);  \coordinate (V6) at (-\s,-\a);
  \coordinate (V7) at ( \s,-\a);  \coordinate (V8) at ( \a,-\s);

  % Bounding box B of O_p: shade it, then cover O_p so only B \ R stays shaded
  \fill[blue!12] (-\a,-\a) rectangle (\a,\a);
  \fill[white] (V1) -- (V2) -- (V3) -- (V4) -- (V5) -- (V6) -- (V7) -- (V8) -- cycle;
  % The corner of B \ R where the diagonal facet's box bound peaks: darker, so the eye lands there.
  \fill[blue!30] (V1) -- (\a,\a) -- (V2) -- cycle;

  % Chan & Mitra's guard rho <= 100 m, over-approximated by O_p
  % \draw[guard] (0,0) circle (5);

  % Bounding box outline (under the hull so coincident edges show both)
  \draw[bbox] (-\a,-\a) rectangle (\a,\a);

  % ---- LP concretization (delete this block for the plain version) ----
  % Diagonal velocity facet (rdvh_vel5, a = e(225 deg)): the CROWN plane's
  % position gradient is -0.576 a, i.e. along (1, 1).  l(p) = x + y:
  %   box concretization:  y_hat_0 = 2 a_p      = 10     at (a_p, a_p)
  %   per-tile LPs:        z_1 = sqrt2 a_p      = 7.07   on the facet V1--V2 (tile E);
  %                                                      the LP returns the top vertex V2
  %                        z_2 = a_p + h        = 5.75   at ( h, a_p)        (tile N)
  %                        z_3 = 0                     at (-h, h)          (tile W)
  %                        z_4 = a_p - h        = 4.25   at (a_p, -h)        (tile S)
  %   y = max_i z_i = z_1 (the whole facet: the degenerate case of the LP)
  \begin{scope}
    \clip (-\a,-\a) rectangle (\a,\a);
    \draw[level] (1.5,5.571) -- (5.571,1.5);          % level set of l through V1--V2 clipped to square
  \end{scope}
  \draw[purple!70!black, -{Stealth[length=5pt]}, line width=0.9pt] (2.6,2.6) -- (3.15,3.15);
  \node[purple!70!black, font=\small, below right, inner sep=1pt] at (2.75,2.65) {$\nabla\overline{\mathbf{A}}$};
  % Axis facet (rdvh_vel4, a = e(180 deg)) for contrast: gradient -0.576 a along +x, l'(p) = x.
  % Box maximizers: the whole face x = a_p of B, which contains the facet V8--V1, so y_hat_0' = y'.
  \draw[axisc, -{Stealth[length=5pt]}, line width=0.9pt] (2.3,-3.3) -- (3.1,-3.3);
  \node[axisc, font=\small, left, inner sep=2pt] at (2.3,-3.3) {$\nabla\overline{\mathbf{A}^\prime}$};
  % ---------------------------------------------------------------------

  % Outer boundary of R: the octagon
  \draw[hull] (V1) -- (V2) -- (V3) -- (V4) -- (V5) -- (V6) -- (V7) -- (V8) -- cycle;

  % Tile boundaries (K = 4 disjoint pinwheel tiles, polyreach/spre.py docking_hole):
   % P1 = E {x >= h, y >= -h}, P2 = N {y >= h, x <= h}, P3 = W {x <= -h, y <= h}, P4 = S {y <= -h, x >= -h}
  \draw[poly] ( \h, \h) -- ( \h, \a);   % P1 | P2
  \draw[poly] (-\h, \h) -- (-\a, \h);   % P2 | P3
  \draw[poly] (-\h,-\h) -- (-\h,-\a);   % P3 | P4
  \draw[poly] ( \h,-\h) -- ( \a,-\h);   % P4 | P1

  % Avoid set: ARCH's docking (collision) box |x|, |y| < 1 m
  \filldraw[pillar] (-\h,-\h) rectangle (\h,\h);

  % ---- LP concretization markers (part of the block above) ----
  \node[single] at (\a,\a) {};
  \node[blue!70!black, font=\small, above right, inner sep=1.5pt] at (\a,\a) {$\hat{y}_0$};
  \draw[orange!80!black, line width=2.4pt] (V1) -- (V2);
  \draw[axisc, line width=2.4pt] (V8) -- (V1);
  % Looseness of box concretization: from the facet (y) out to the corner (hat y_0).
  \draw[blue!70!black, {Stealth[length=4pt]}-{Stealth[length=4pt]}, line width=0.9pt]
    (3.66,3.66) -- (4.86,4.86);
  % label inside the square, next to the green facet
  \node[axisc, font=\small, left, inner sep=2pt] at (4.85,-1.75) {$\hat{y}^\prime = y^\prime$};

  % z1: the LP returns a vertex of the optimal facet, here the top vertex V2
  \node[piece] at (V2) {};
  \node[orange!80!black, font=\small, above, inner sep=2pt] at (V2) {$y = z_1$};

  \node[piece] at (\h,\a) {};
  \node[font=\small, below left, inner sep=2pt] at (\h,\a) {$z_2$};
  \node[piece] at (-\h,\h) {};
  \node[font=\small, above left, inner sep=1.5pt] at (-\h,\h) {$z_3$};
  \node[piece] at (\a,-\h) {};
  \node[font=\small, below left, inner sep=2pt] at (\a,-\h) {$z_4$};
  % -------------------------------------------------------------

  % Labels
  \node at ( 2.6, 0.9) {$\mathcal{P}_1$};
  \node at (-0.9, 2.6) {$\mathcal{P}_2$};
  \node at (-2.6,-0.9) {$\mathcal{P}_3$};
  \node at ( 0.9,-2.6) {$\mathcal{P}_4$};
  \node[blue!70!black] at (-4.45,-4.45) {$\mathcal{B}_0$};
  \node[red!80!black, font=\small, below right, inner sep=1pt] at (\h,-\h) {dock};
  \node[right] at (5.6,-3.6) {$\mathcal{R}$};
  \draw[black!60, -{Stealth[length=4pt]}] (-6.0,-5.8) -- (-5.0,-5.8) node[right, font=\small] {$x$};
  \draw[black!60, -{Stealth[length=4pt]}] (-6.0,-5.8) -- (-6.0,-4.8) node[above, font=\small] {$y$};
\end{tikzpicture}

%% file: results.tex
\begin{tabular}{@{}ll rr rr@{}}
\toprule
 & & \multicolumn{2}{c}{branches} & \multicolumn{2}{c}{time (s)} \\
\cmidrule(lr){3-4} \cmidrule(lr){5-6}
System & Verifier & total & max & $T_1$ & $T_\infty$ \\
\midrule
\multirow{4}{*}{\shortstack[l]{SPRE\\[1pt]\footnotesize 8 (8)}} & C\&V, $\mathcal{R}$ & 28 & 6 & 6.6 & 1.0 \\
 & C\&V, GPU & 21 & -- & 7.5 & -- \\
\cmidrule(l){2-6}
 & C\&V, Fit-LRP & \textbf{8} & \textbf{1} & 5.1 & 0.7 \\
 & Fit-LRP BaB & \textbf{8} & \textbf{1} & \textbf{3.0} & \textbf{0.5} \\
\midrule
\multirow{4}{*}{\shortstack[l]{TLPRA\\[1pt]\footnotesize 3{,}068 (756)}} & C\&V, $\mathcal{R}$ & 303{,}942 & 36{,}372 & 9{,}701 & 103 \\
 & C\&V, GPU & 56.5M & -- & 27{,}205 & -- \\
\cmidrule(l){2-6}
 & C\&V, Fit-LRP & 167{,}795 & 2{,}198 & 8{,}806 & 12 \\
 & Fit-LRP BaB & \textbf{135{,}290} & \textbf{1{,}853} & \textbf{8{,}253} & \textbf{8.9} \\
\midrule
\multirow{4}{*}{\shortstack[l]{MSIS\\[1pt]\footnotesize 2{,}710 (218)}} & C\&V, $\mathcal{R}$ & 474{,}538 & 42{,}543 & 6{,}078 & 71 \\
 & C\&V, GPU & 127.3M & -- & 25{,}665 & -- \\
\cmidrule(l){2-6}
 & C\&V, Fit-LRP & 366{,}869 & 38{,}125 & 5{,}791 & 63 \\
 & Fit-LRP BaB & \textbf{38{,}356} & \textbf{909} & \textbf{4{,}178} & \textbf{4.8} \\
\bottomrule
\end{tabular}